# Asynchronous and Segmented Bidirectional Encoding for NMT

Jingpu Yang[a], Zehua Han[a], Mengyu Xiang[a], Helin Wang[a], Yuxiao Huang[a], Miao Fang[a,b*]
[a]Northeastern University, Shenyang, China
[b]Olimei Company, Guangzhou, China

## ABSTRACT

With the rapid advancement of Neural Machine Translation (NMT), enhancing translation efficiency and quality has become a focal point of research. Despite the commendable performance of general models such as the Transformer in various aspects, they still fall short in processing long sentences and fully leveraging bidirectional contextual information. This paper introduces an improved model based on the Transformer, implementing an asynchronous and segmented bidirectional decoding strategy aimed at elevating translation efficiency and accuracy. Compared to traditional unidirectional translations from left-to-right or right-to-left, our method demonstrates heightened efficiency and improved translation quality, particularly in handling long sentences. Experimental results on the IWSLT2017 dataset confirm the effectiveness of our approach in accelerating translation and increasing accuracy, especially surpassing traditional unidirectional strategies in long sentence translation. Furthermore, this study analyzes the impact of sentence length on decoding outcomes and explores the model's performance in various scenarios. The findings of this research not only provide an effective encoding strategy for the NMT field but also pave new avenues and directions for future studies.

## 1. INTRODUCTION

In today's digital and globalized era, the demand for language translation is growing exponentially, particularly in the context of the information explosion and frequent cross-cultural communication, making efficient and accurate machine translation systems increasingly vital. Neural Machine Translation (NMT)[2][3], as a cutting-edge form of automatic translation technology, has made significant strides in recent years. The advent of the Transformer model [1], in particular, has greatly propelled the development of NMT technology, achieving breakthroughs in handling complex language patterns and enhancing translation quality. However, despite these achievements, existing NMT systems still face challenges in processing long sentences and fully leveraging bidirectional contextual information [4].

The Transformer model [1], with its incorporation of attention mechanisms [6] and an encoder-decoder framework [7], has made significant advances in the field of NMT [5]. However, the translation of long sentences remains a considerable challenge [8], primarily due to the complex grammatical structures and deep semantic relationships they entail. Moreover, most existing NMT systems employ a unidirectional decoding strategy, whether left-to-right or right-to-left [9], which limits their capacity to fully understand the context of entire sentences [10]. Consequently, developing a translation model that can effectively process long sentences and fully utilize bidirectional contextual information is crucial for enhancing the overall performance of NMT systems.

To address these challenges, we developed an enhanced Transformer-based model, implementing an asynchronous and segmented bidirectional decoding strategy. By incorporating reverse decoding and deepening the residual network layers, we significantly improved the efficiency and accuracy of long sentence translation. Moreover, the model more comprehensively utilizes bidirectional contextual information, thereby further enhancing translation quality. Experimental results on the IWSLT2017 dataset demonstrate that our model outperforms traditional unidirectional translation methods in terms of efficiency and quality, especially in handling long sentences. We also explored the impact of sentence length on translation outcomes and analyzed the model's performance across different scenarios. This approach offers us an efficient machine translation solution, surpassing the original unidirectional decoding model in both translation speed and accuracy.

* Corresponding author (fangmiao@neuq.edu.cn)

## 2. RELATED WORK

### 2.1 Neural Machine Translation

Neural Machine Translation (NMT) utilizes deep learning techniques to facilitate language translation [11]. Since the introduction of an encoder-decoder framework with an attention mechanism, NMT has emerged as the predominant technology in the field of machine translation [14]. Unlike traditional rule-based or statistical translation methods, NMT learns the mapping from source to target language directly through neural networks. This breakthrough lies in its ability to handle long-range dependencies in language more effectively, resulting in smoother and more natural translations. The Transformer model [1], introduced by Vaswani et al. in 2017, has further advanced NMT with its unique self-attention mechanism and parallel processing capabilities. Departing from traditional Recurrent Neural Network (RNN) [15] and Convolutional Neural Network (CNN) [16] architectures, the Transformer employs a multi-head attention mechanism to efficiently capture dependencies between different positions, thereby significantly enhancing performance across various NMT tasks.

Despite significant progress in translation quality and efficiency, Neural Machine Translation (NMT) still encounters substantial difficulties and challenges when dealing with specific types of texts, such as long sentences and low-resource languages [17]. The complexity of translating long sentences lies in maintaining the integrity of sentence structure and coherence of context. Additionally, NMT systems typically require extensive bilingual corpora for training, which poses a limitation in the context of low-resource languages. Consequently, researchers continue to explore new architectures and optimization strategies to further enhance the performance and adaptability of NMT systems. We have effectively mitigated these issues through the implementation of synchronized bidirectional encoding in both forward and reverse directions.

### 2.2 Neural Machine Translation of Long Sentences

In the realm of NMT, handling long sentences poses a significant challenge. Long sentences typically involve complex grammatical structures and rich semantic information, presenting a challenge to the memory and processing capabilities of neural network models. Traditional NMT models based on RNN, especially Long Short-Term Memory (LSTM) networks [18], have achieved certain success in managing long-range dependencies, but still grapple with issues like vanishing or exploding gradients [20]. While the Transformer model captures long-distance dependencies more effectively, its limitations in translating extremely long sentences remain. Consequently, researchers have been exploring various methods to optimize the translation of long sentences, such as sentence segmentation techniques and hierarchical attention mechanisms. Cho et al. (2014) proposed a segmentation-based approach for long sentences, where a long sentence is divided into shorter segments, translated individually, and then the translations are merged [19]. However, this approach does not fundamentally address the issue of model forgetting earlier tasks in long sentence translation.

### 2.3 Combining Forward Encoding and Reverse Encoding

In the field of NMT, the integration of forward and reverse encoding has become a significant research direction aimed at enhancing the performance and accuracy of translation models. This approach, by concurrently considering the forward and backward information of the source sentence, allows for a more comprehensive capture of the sentence's semantic and structural characteristics. Forward encoding captures the sequential information from the start to the end of the sentence, while reverse encoding focuses on the reverse order, from the end to the start. This bidirectional encoding strategy aids in a better overall understanding of the sentence's semantics. Sutskever, Vinyals, and Le (2014) were among the first to propose the concept of combining forward and reverse encoding within a sequence-to-sequence learning framework [21]. They discovered that this method of integrating bidirectional information significantly improved the performance of machine translation. Subsequent researchers have implemented similar combined strategies in various NMT models, such as using Bidirectional Recurrent Neural Networks (BiRNN) [22] or Bidirectional Long Short-Term Memory networks (BiLSTM) [23] to process sentence representations in both forward and reverse directions. Furthermore, recent studies have demonstrated the effective application of combining forward and reverse encoding strategies within the Transformer model [24]. By modifying the self-attention mechanism to account for both forward and backward contextual information, we have found that it is possible to further enhance the quality and accuracy of translations without compromising speed. This indicates that the integration of forward and reverse encoding is an effective method to improve the performance of NMT systems, whether in traditional RNN architectures or in the latest Transformer models.

## 3. ASYNCHRONOUS AND SEGMENTED BIDIRECTIONAL ENCODING

In this study, we introduce an innovative NMT model that integrates both forward and reverse translation strategies. This integrated approach effectively addresses some of the challenges encountered by traditional models, particularly in the translation of long sentences. Typically, in standard forward translation, models may underperform in the latter parts of the sequence, while reverse translation might yield poorer results in the initial segments. Our model, by concurrently considering the outcomes of both forward and reverse translations, ensures a balanced and high-quality translation throughout the sentence.

To tackle the common issues of gradient vanishing and explosion in the training of deep neural networks [24], our model incorporates additional layers of residual networks. During the decoding process, the forward translation path includes two layers of residual networks, whereas the reverse translation path is equipped with one layer. This design enables the model to more effectively capture and integrate information from both directions, thereby facilitating the acquisition of a richer knowledge base. The introduction of this deep structure not only enhances the learning capacity of the model but also significantly improves the accuracy of translation. The overall model is shown in Figure1.

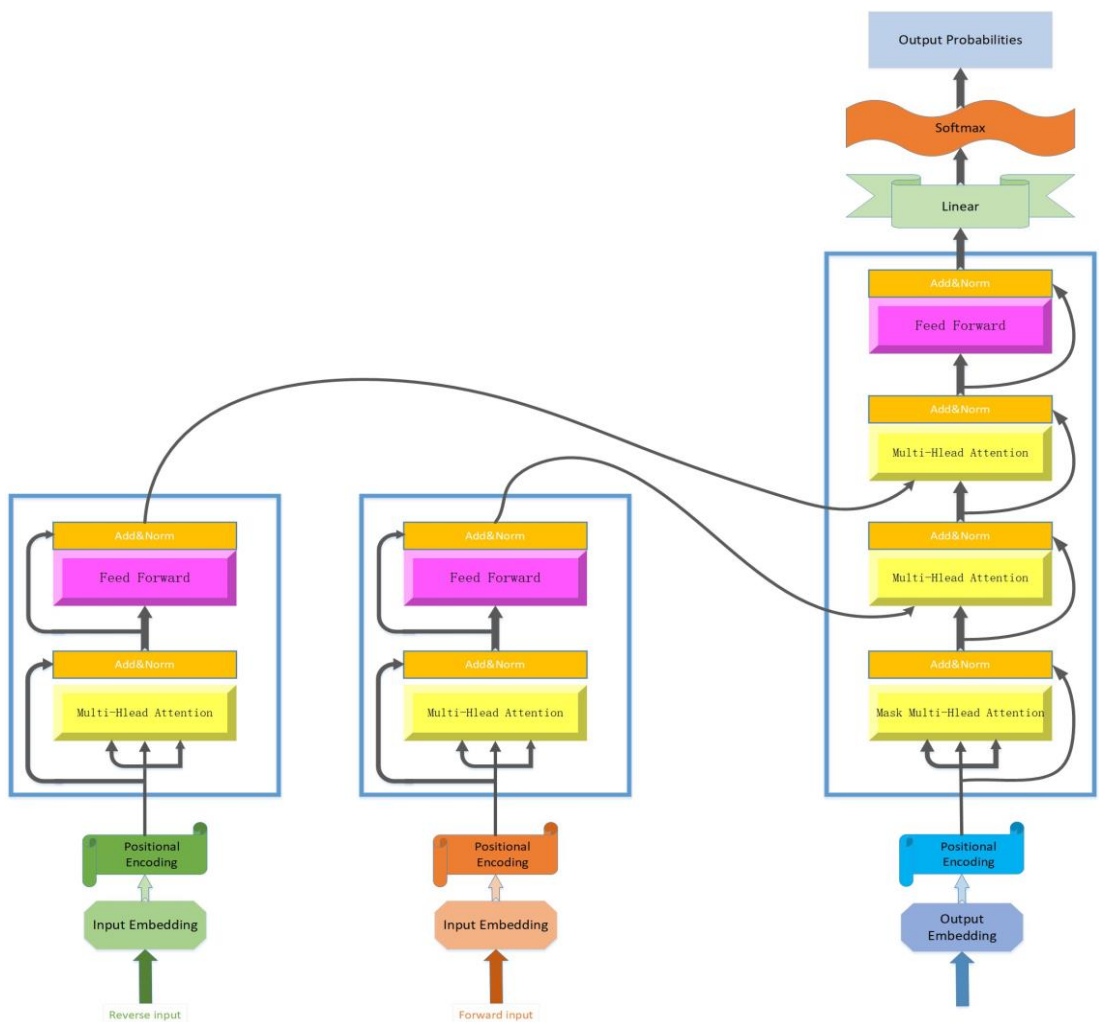

Figure 1. Visualization of the our model. Green represents reverse input and orange represents forward input.

## 4. EXPERIMENTS

### 4.1 Evaluation Setting

In this study, the evaluation was conducted using the IWSLT2017 English-to-German dataset [25], a widely utilized standard dataset in the field of Neural Machine Translation for measuring the performance of various translation models. The parameters of our experimental model were aligned with those of the original Transformer model to ensure the validity and comparability of our assessment. To achieve optimal training results, we set a training duration of 100 epochs and implemented an early stopping mechanism to prevent overfitting. This mechanism was based on the model's performance on the validation set, and training was prematurely terminated when there was no further improvement in model performance.

For baseline comparison, we utilized two conventional Transformer models, configured for translation in the left-to-right and right-to-left directions, respectively. The setup of these baseline models allowed us to accurately gauge the performance differences between our proposed model and standard methodologies. The primary criterion for evaluating model performance was the sentence-level BLEU (sentence blue) score [26], a widely used and effective metric for assessing the accuracy of machine translation. The final results were determined by the highest sentence-level BLEU scores recorded across the 100 training epochs, ensuring that our evaluation reflected the best performance of the model throughout the training process. This evaluation setup was designed to comprehensively and accurately assess the efficacy and advantages of our improved Transformer model in handling complex translation tasks.

### 4.2 Results and Evaluation of the Model on the Dataset

In this study, we compared our proposed model with two baseline Transformer models oriented in left-to-right (l2r) and right-to-left (r2l) translation directions. The experimental results indicated that our model achieved a sentence-level BLEU score of 14.836, outperforming the standard left-to-right (Transformer l2r) and right-to-left (Transformer r2l) Transformer models, which scored 13.760 and 13.102, respectively. These findings demonstrate a significant improvement in translation quality by our model compared to conventional unidirectional translation methods. Particularly in the translation of long sentences, the higher BLEU score of our model reflects its enhanced accuracy and fluency. These results validate the effectiveness of our model's integrated approach of combining forward and reverse translation strategies, and its potential to elevate translation performance. The specific results are shown in Table1

| model | Our model | Transformer l2r | Transformer r2l |
|---|---|---|---|
| Sentence blue score | 14.836 | 13.760 | 13.102 |

Table 1. The highest sentence-level BLEU scores were obtained for each model after 100 training epochs, and our model was compared with two baseline Transformer models (left-to-right and right-to-left translation) on the IWSLT2017 English-to-German dataset.

We categorized the sentences in the dataset by length and computed the accuracy of the model across different sentence lengths. As illustrated in Figure2, our model consistently demonstrated higher sentence-level BLEU scores across various sentence lengths, with scores progressively increasing from an initial 12.5 to 15.8, indicating a significant and stable upward trend. In contrast, the conventional Transformer models in left-to-right (l2r) and right-to-left (r2l) configurations exhibited a decline in performance, particularly in the translation of longer sentences. For instance, the Transformer l2r model's score decreased from a peak of 13.5 to 11.7, while the Transformer r2l model dropped from 13.9 to 11.1. These results highlight our model's remarkable stability and efficiency in translating long sentences, especially in handling complex long sentence translation tasks. Our model, by integrating forward and reverse translation strategies and incorporating additional layers of residual networks, significantly enhanced the learning capability and thereby improved the accuracy of translation. In contrast, conventional unidirectional Transformer models encountered performance bottlenecks when faced with complex long sentences. These findings comprehensively validate the effectiveness of our proposed model in enhancing the accuracy of long sentence translations.

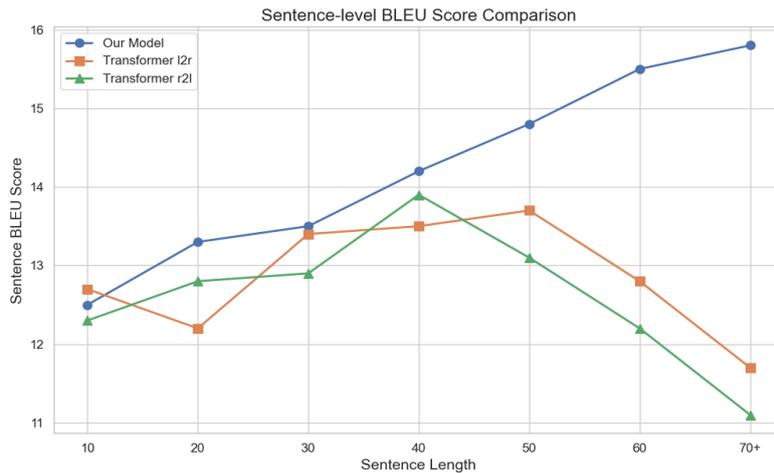

Figure 2. The sentences in the dataset were categorized by length, and the Sentence BLEU scores of the model were calculated and compared across different sentence lengths.

## 5. CONCLUSION

In this study, we introduced an innovative Neural Machine Translation (NMT) model that integrates forward and reverse translation strategies, and it was comprehensively evaluated on the IWSLT2017 English-to-German dataset. After undergoing 100 training epochs, our model demonstrated superior translation performance across various sentence lengths, particularly excelling in the translation of long sentences compared to traditional unidirectional Transformer models. In contrast to standard left-to-right and right-to-left Transformer models, our model not only showed higher stability and efficiency in translating long sentences but also achieved significant improvements in overall sentence-level BLEU scores.By incorporating additional layers of residual networks, our model effectively addressed the issues of gradient vanishing and explosion commonly encountered in the training of deep neural networks, thereby enhancing learning capabilities and translation accuracy. These experimental results validate the effectiveness of our model in improving the quality of long sentence translations and maintaining training stability, offering a new direction for the development of future NMT models. In summary, our research not only enhances the accuracy of machine translation but also provides an effective solution for handling complex translation tasks, particularly in translating long sentences. Moving forward, we look forward to further validating and optimizing our model on a broader range of language pairs and larger datasets.